# An Improved Transformer-based Model for Detecting Phishing, Spam, and Ham – A Large Language Model Approach

**Suhaima Jamal**
*Dept. of Information Technology*
*Georgia Southern University*
Statesboro GA, USA
sj14077@georgiasouthern.edu

**Hayden Wimmer**
*Dept. of Information Technology*
*Georgia Southern University*
Statesboro GA, USA
hwimmer@georgiasouthern.edu

**Iqbal H. Sarker**
*Security Research Institute,*
*Edith Cowan University,*
*Perth, WA-6027, Australia*
m.sarker@ecu.edu.au

## Abstract

Phishing and spam detection is long standing challenge that has been the subject of much academic research. Large Language Models (LLM) have vast potential to transform society and provide new and innovative approaches to solve well-established challenges. Phishing and spam have caused financial hardships and lost time and resources to email users all over the world and frequently serve as an entry point for ransomware threat actors. While detection approaches exist, especially heuristic-based approaches, LLMs offer the potential to venture into a new unexplored area for understanding and solving this challenge. LLMs have rapidly altered the landscape from business, consumers, and throughout academia and demonstrate transformational potential for the potential of society. Based on this, applying these new and innovative approaches to email detection is a rational next step in academic research. In this work, we present IPSDM, our model based on fine-tuning the BERT family of models to specifically detect phishing and spam email. We demonstrate our fine-tuned version, IPSDM, is able to better classify emails in both unbalanced and balanced datasets. This work serves as an important first step towards employing LLMs to improve the security of our information systems.

**Keywords:** Large language model (LLM), Phishing, Spam, Artificial Intelligence, Cyber Security, Fine Tuning, DistilBERT, RobBERTA

## 1. Introduction

Phishing and spam are a continual concern that continues to plague users and cost economies the world over financial resources. Individuals and businesses suffer from fiscal loss and financial setback due to such issues caused by spam and phishing attacks. This kind of fraudulent endeavors attempt to deceive individual into revealing sensitive and confidential information, such as financial details or login credentials etc. While Artificial Intelligence (AI) approaches have attempted to assuage these issues [1], heuristic-based systems continue to dominate. Radical new approaches have emerged due to advances in technology and increased research investment by both the public and private sector. The recent advancements in AI-based solutions have led to the development of innovative and unconventional strategies to combat spam and phishing tactics such as demonstrated by Anand, et al. [2].

Transformer-based models possess a revolutionary impact on the development of spam and phishing classification models while processing, understanding, and interpreting the text data inputs. For email-based datasets, such models are constantly evolving with several regularization methods. Furthermore, attention-based mechanisms in transformer allows model interpretability and makes it easier in understanding certain classification decision [3]. Large Language Models (LLMs), made famous by Open AI's ChatGPT, have emerged triumphant in solving new problems while being adapted to well-established challenges such as phishing and spam. Open AI's ChatGPT runs on its GPT engine and has been able to make large strides in consumer and business adoption [4]. The most famous competing LLMs are available from a plethora of vendors such as Google, Meta, and MIT to name a few and the emergence of competing LLMs such as Llama and Bert have been open sourced thereby fueling research and development from large institutions all the way down to the consumer. While these models are available for download, the ability to run pre-trained models such as BERT is still in nascent stages with more consumers having access to local GPU technology as well as organizations like Google with Collaboratory and Hugging Face with its transformer's library and model hosting.

LLMs are general in nature and pre-trained by the creators and published for commercial and non-commercial licenses. There are a multitude of inputs to train a LLM such as web scraping, document corpus, and even text sources such as email and transcribed books, discussions, or speeches. While LLMs perform well on general tasks, they can be fine-tuned to improve their performance on more specific tasks. One such example is FinBERT [5] where BERT (Bidirectional Encoder Representations from Transformers) was trained on financial specific documents and is able to better respond to use prompts on finance. Other such advances are in progress for medical data to aid in both physician decision making and end-user queries. BERT employs a self-attention mechanism which enables the model to capture both contextual information and dependencies among words in any text sequence. Through the self-attention method, the weights of relevant important words are calculated. Attention scores are measured for all words or input tokens and passed through SoftMax function. A rich contextual embedding can be generated by BERT based models which allow to excel in several natural language understanding tasks.

Within the family of BERT-based models, DistilBERT and RoBERTA are two significant variants and have been used for tasks such as fake news [6] or to make predictions via Twitter data [7]. Both models are built based on transformer architecture while excelling in NLP processing tasks. DistilBERT is designed for reducing the number of parameters making it faster and smaller version of BERT. Whereas Roberta is considered as more optimized and robust version. In this work, we present an Improved Phishing and Spam Detection Model (IPSDM), a custom trained and fine-tuned version of DistilBERT and RoBERTA. We fine-tuned these models specifically on phishing, spam, and ham data from multiple sources. We demonstrate that our fine-tuned IPSDM outperforms basic BERT and RoBERTA on both imbalanced and balanced datasets of phishing, spam, and ham.

The rest of the paper is outlined as follows; section 2 describes the related current-state-of arts research works. Section 3 provides a detailed explanation of the proposed model's framework and methodology. In section 4, the experimental outcomes and results are broadened. Furthermore, section 5 encompasses an elaborated discussion of the results. Finally, section 6 holds the concluding remarks and future prospects of this work.

## 2. Literature Survey

### 2.1 Machine Learning and Deep Learning based Methods

Numerous machine learning and deep learning-based spam email detection and classification applications have been carried out over the past few decades by many researchers. In such studies [8-13], authors have proposed, reviewed, and evaluated spam filtering models where the classification models are based on traditional machine learning algorithms, i.e., Naïve Bayes, Random Forest, SMV mostly. Govil, et al. [8] have created a dictionary, named "stopwards" for removing the helping verbs from email. Then, the algorithm is executed for checking the possibility of being spam or not. A machine learning classifier, Naïve Bayes has been applied for the identification purpose where non-spam emails were classified as spam, 1 and non-spam, 0 [8].

Similarly, Chen, et al. [9] have evaluated machine learning algorithms for detecting spam tweets. A large dataset containing around 600 million public tweets have been collected first. Later, Trend Micro's Web Reputation System was applied for labeling the spam emails. Experiments on different data sizes revealed that TP rate is increased from 78% to 85% following KNN and 70% to 75% following Random Forest classifier. Another potential finding is the classifier could detect continuously sampled spam tweets better than randomly selected tweets [9]. In the similar context of Twitter spam detection, Wu, et al. [14] introduced a WordVector Training based model with a classification accuracy of around 80%. Average 30% higher F-measures have been achieved in this work compared to other existing models.

Moreover, Guzella and Caminhas [12] reviewed the textual and image-based spam email filtering approaches focusing on designing new filters. Most common method selecting the feature is information gain and this way of collecting features might increase accuracy. In terms of datasets, SpamAssassian and LingSpam are considered as the most popular ones, whereas TREC corpora can produce more realistic online setting. Moreover, Chetty, et al. [15] proposed a deep learning-based model combining Word Embedding and Neural Network aiming to detect spams from various text documents. Naïve Bayes model is considered as the baseline model for comparing with the deep learning model. Datasets were collected from UCI machine learning repository for developing the models. For SMS dataset, the highest performance (accuracy 98.7%) is achieved from the combined model of Word Embedding and neural network. Apart from the supervised learning approaches, there are numerous works on unsupervised modeling as well [16-21]. Utilizing Modified Density-Based Spatial Clustering of Applications with Noise (M-DBSCAN), 97.848% accuracy has been obtained by Manaa, et al. [17]. An online unsupervised spam detection scheme, SpamCampaignAssassin (SCA) could detect around 92.4% spam for DEPT trace email dataset [16].

## 2.2 Transformer model-based approaches

The research works and literature landscape on transformer-based methods are relatively limited. The domain of fine-tuned transformers or attention mechanism techniques for identifying spam emails is still an emerging new field. Related to this specific area, Yaseen [22] has introduced an effective word embedding technique for spam classification. Pre-trained transformer, BERT is fine tuned to detect the spam emails from non-spam emails. Deep Neural Network with BiLSTM is considered as a baseline model to compare the model. Two open-source datasets from UCI machine learning repository and Kaggle have been employed to train and test the model. The proposed model could achieve 98.67% classification accuracy. Similarly, Liu, et al. [23] have developed and evaluated a modified spam detector transformer using the publicly available datasets, Spam Collection v.1 and UtkMI's Twitter Spam Detection Competition dataset. This model could obtain 98.92% accuracy with a recall and F1 scores rate respectively, 0.9451 and 0.9613.

Furthermore, Guo, et al. [24] and Tida and Hsu [25] focused on BERT models implying the significance of self-attention mechanism. Guo, et al. [24] utilized two public datasets, Enron [26] and simple spam email classifier dataset from Kaggle for classifying ham or spam emails using pre-trained BERT model. Similarly, an Universal Spam Detection Model (USDM) has been developed and tested using four publicly available datasets which are Ling-spam dataset [27], spam text dataset from Kaggle, Enron dataset and spam assassin dataset. This model has gained overall accuracy of 97% with 0.96 F1 score [25] . Moreover, for detecting phishing URL, researchers worked on fine tuning BERT based models [28, 29]. Wang, et al. [28] have scrapped 2.19 million pieces of URL data from PhishTank while pre-training PhishBERT model. This model exhibited 92% accuracy in detecting phishing URLs. Similarly, Maneriker, et al. [29] fine-tuned BERT and RoBERTa models and proposed a URLTran transformer. Microsoft Edge and Internet Explorer browsing telemetry data have been employed for training, testing, and validating purpose. Down sampling method is applied for balancing the datasets where the final training dataset had 77,870 URLs. The final models had a True Positive Rate (TPR), 86.80% compared to the baseline models URL-Net [30] and Texception [31].

## 3. Methodology

In this paper, transformer-based self-attention mechanism models are explored with an aim to improving the pre-trained baseline BERT models. Our collected and prepared dataset is used for developing and comparing models in two different settings. 1) DistilBERT and RoBERTA were pretrained using both imbalanced and balanced phishing-ham-spam dataset, and 2) the base models' training process has been improved through applying optimization and fine-tuning mechanism. We named our proposed model as Improved Phishing Spam Detection Model (IPSDM). This model's classification performance is compared with the baseline models (DistilBERT and RoBERTA). At the end of the experiment, IPSDM exhibited substantial improvement in performance both for balanced and imbalanced scenarios compared to baseline models while detecting phishing and spam emails and texts. The top-level methodology of this research is presented in figure 1. Later, the breakdown of detailed flow diagram of model optimization and fine-tuning are illustrated in figure 6 and 9 of section 2.4.

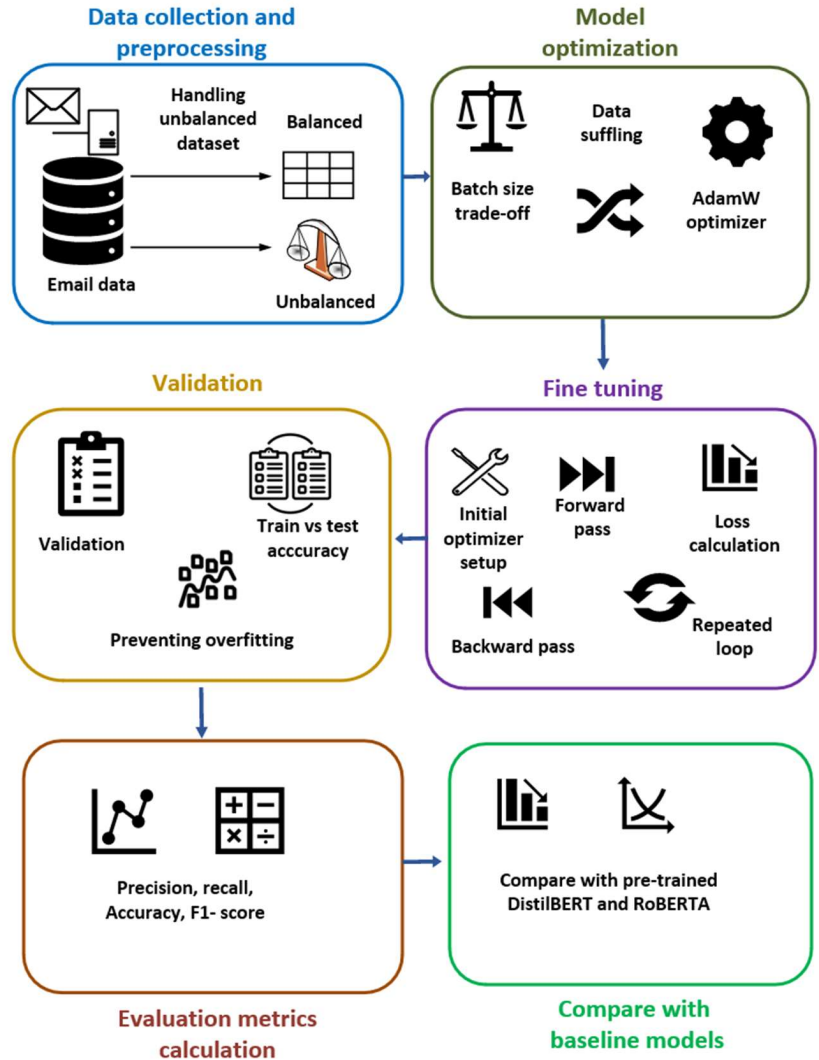

Figure 1. Overall methodology

## 3.1 Data Collection and Preparation

The data for training, testing, and validating this experiment is developed by concatenating two opensource data sources [32, 33]. One dataset has ham and spam emails which is merged with another phishing email dataset. The concatenated dataset has 747 spams, 189 phishing and 4825 ham samples which is highly imbalanced. This has been further resampled following adaptive synthetic sampling (ADASYN) technique where minor classes (ham and spam) are oversampled by generating synthetic samples with a focus on difficult-to-learn instances. This process reduces the bias towards the majority class making the overall predictive model more accurate and efficient. This versatile technique of sampling assists in mitigating the risk of overfitting as well. Figures 2 presents the feature distribution before and after sampling. Figure 3 shows a snapshot of final dataset.

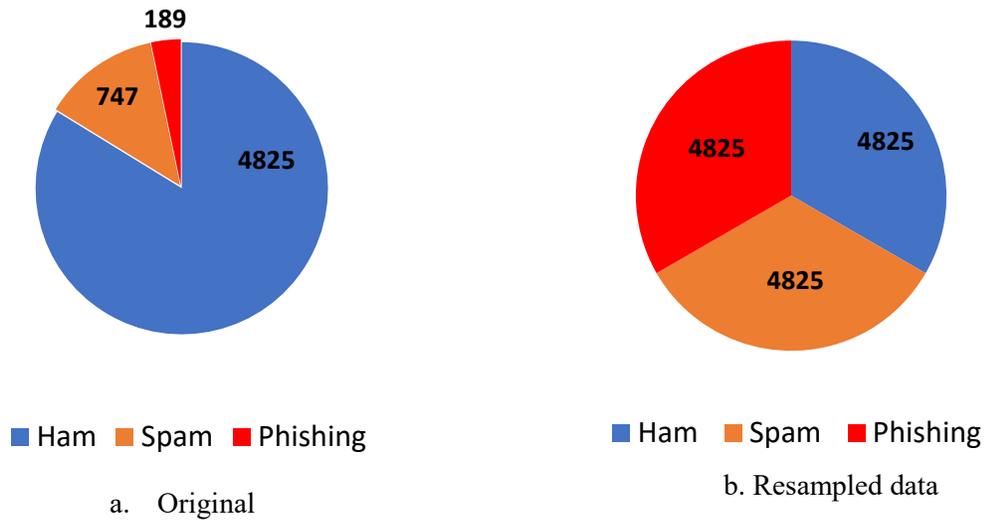

a. Original

b. Resampled data

2. Feature distribution

| Email | Category |
|---|---|
| Go until jurong point, crazy.. Available only in bugis n great world la e buffet... Cine there got amore wat... | ham |
| Ok lar... Joking wif u oni... | ham |
| U dun say so early hor... U c already then say... | ham |
| Shop till u Drop, IS IT YOU, either 10K, 5K, √•¬£500 Cash or √ •¬£100 Travel voucher, Call now, 09064011000. NTT PO Box CR01327BT fixedline Cost 150ppm mobile vary | spam |
| Nah I dont think he goes to usf, he lives around here though | ham |
| refund confirmation | phishing |
| Even my brother is not like to speak with me. They treat me like aids patent. | ham |

Figure 3. A snapshot of dataset overview

### 3.2 Data Splitting

The overall dataset is split into 80% (training set) and 20% (testing set). Later, from the 80% set, 60% kept for training and 20% for validation. This 20% validation set is used after the completion of each training epoch which aids in identifying the optimal model performance. It is an integral part of the development process that ensures the model's effectiveness on unseen data identification and prediction.

### 3.3 Model Selection

#### A. DistilBERT

DistilBERT is a derivation of Bidirectional Encoder Representations from Transformers (BERT) which is a transformer-based model pre-trained for developing natural language processing tasks. The idea here is to compress the original model for making it more computationally efficient and faster [34]. The models can be further finetuned for any specific downstream tasks on any customized dataset. DitilBERT model achieves the compression by mimicking a teacher-student model where the customized model is trained. The input tokens are the raw text inputs that need to be preprocessed. The tokenizer uses a vocabulary to tokenize the input words into sub-words. Later, the tokenized inputs are mapped to numerical embedding. The relationships between the words are captured through the attention layer. This attention mechanism works by calculating the attention score between tokens inside a sequence allowing the model to focus more on the significant relevant words than the irrelevant ones. The pooling section indicates the entire input sequence having a fixed representation. The classifier head can be modified for any specific task and the final prediction layer predicts the corresponding model output. For our case, this is detecting spam/ham/phishing emails.

### B. RoBERTA

A Robustly Optimized BERT Pretraining Approach (RoBERTA) is an extended version of the transformer-based model, BERT where model can operate on large batch size and train longer sequence. The pretraining process follows improved bidirectional context-oriented mechanism while learning the masked-out tokens for longer sequences [35]. The architecture is similar as DistilBERT having transformer encoder layers with multi-head attention mechanisms. However, model has a byte-level tokenizer which is different than BERT. The dynamic masking works at different epochs and uses BPE as a subunit, not as characters. RoBERTA receives tokens as inputs and a tokenizer preprocess these. It passes through encoding, pooling, decoding and attention mechanism. The basic architecture of DistilBERT and RoBERTA model is similar which is illustrated in figure 4.

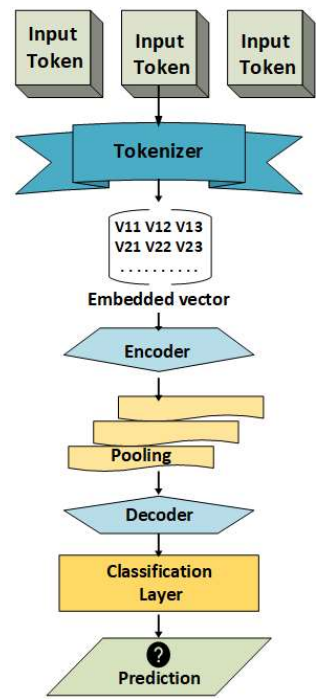

Figure 4. Basic architecture of DistilBERT and RoBERTA

## 3.4 Improving the Training Process

Employing the phishing-ham-spam dataset, base models of DistilBERT and RoBERTA have been measured first. We aim to improve the model performance and efficiency through optimization, i.e., learning rate scheduling, adjusting batch size, sequence length and loss function, hyper parameter tuning, early stopping and fine tuning.

Necessary measures have been taken to handle overfitting issue. At the end of the process, it has been demonstrated that the final achieved accuracy is not affected by overfitting. This proposed methodology is also employed on imbalanced dataset which was collected initially. A noteworthy improvement is observed while developing models with imbalanced dataset as well.

### A. Model Optimization

As mentioned about data preparation stage in section 2.1, the preprocessed final phishing dataset is tokenized using Hugging Face Transformers tokenizer [36]. A sub-word-based approach is utilized by this tokenizer while breaking down the text into small unit. This allows the model to acknowledge the meaning and context of the words. The pre-trained DistilBERT and RoBERTA models are initialized with their respective pre-trained weight obtained from pre-training process. The batch size is set 32 for training data and 64 for the validation data while trading off between memory consumption and training speed. Training data is shuffled in each epoch ensuring the model's visibility to the different unseen data. This will help prevent overfitting issue.

AdamW (Adam Weight Decay), an efficient optimization algorithm is used here for updating the weights of pre-trained models. This algorithm computes the adaptive learning rate for each parameter by combining exponential moving gradient averages and root mean square gradients [37]. L2 regularization, a weight decay mechanism adds penalty to the loss function which is proportional to magnitude squared weights. This promotes the model to utilize small weights and mitigate overfitting risk by reducing the complexity of the acquired parameters. The model's parameter, Z is initialized with exponential decay rate, $\beta 1, \beta 2$ and $\varepsilon$ with a very small value preventing division by zero. Initially, the first moment, $m_0 = 0\ and$ second moment, $v_0 = 0$. In each iteration, the gradient loss is calculated as below,

$$\text{Gradient loss}, g = \nabla_z L(z)$$

Then, the first moment is updated, $m_i = \beta 1 * m_{i-1} + (1 - \beta 1) * g$

The updated second moment, $v_i = \beta 2 * v_{i-1} + (1 - \beta 2) * g^2$

Later, first and second moment bias get corrected,

$$\widehat{m_i} = \frac{m_i}{1 - \beta 1^i}$$

$$\widehat{v_i} = \frac{v_i}{1 - \beta 2^i}$$

Finally, the parameters are updated using AdamW updating rule,

$$Z_i = Z_{i-1} - \frac{learning\ rate}{\sqrt{\widehat{v_i}} + \varepsilon} . (\widehat{m_i} + weight\ decay * Z_{i-1})$$

This weight decay regularization process assists in controlling the growth of parameters values during the training, mitigating the risk of overfitting.

In the context of loss function, as this is a multiclass classification task, Cross-Entropy Loss is used which combines both SoftMax activation and negative log likelihood into a single loss term. The difference between ground truth label and probability are measured here aiming to minimize the loss during the training process. PyTorch provides cross-entropy loss implementation that handles SoftMax computation and logarithmic computation. For a single training epoch, the loss can be defined as follow,

$$Loss_i = - \sum_{k=1}^{n} Z_i, k * \log(p_{i,k})$$

Here, $Z_i, k$ is the ground-truth label and $p_{i,k}$ is predicted probability made by the model.

The cross-entropy loss for the overall training is the average of individual loss,

$$Loss_{total} = 1/n \sum_{k=1}^{n} Loss_i$$

The models output logits for each of the class which is passed through a SoftMax activation function for converting them into class probability. The predicted probability $p_{i,k}$ is computed as below, where $Z_i, k$ is the produced logit value.

$$p_{i,k} = \frac{e^{Z_{i,k}}}{\sum_{m=1}^{k} e^{Z_{i,m}}}$$

The optimization process diagram is presented in figure 5.

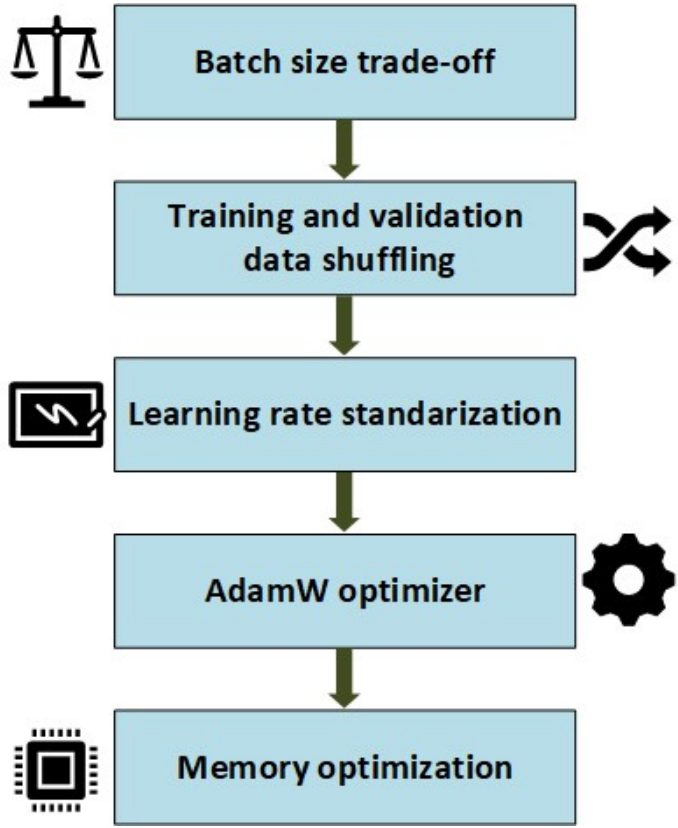

Figure 5. Model optimization

### B. Learning Rate

An ideal learning rate for model optimization and fine tuning depends on several factors, including model architecture, optimization algorithms and the specific task domain. It is a crucial parameter which controls step size during the optimization process. Having a high learning rate might lead the model in unstable mode resulting poor performance for unseen data. Again, lower learning rate can slow down the convergence process. The training process might require more epochs for achieving a good result resulting in higher computational cost. An ideal learning rate graph is presented in figure 6.

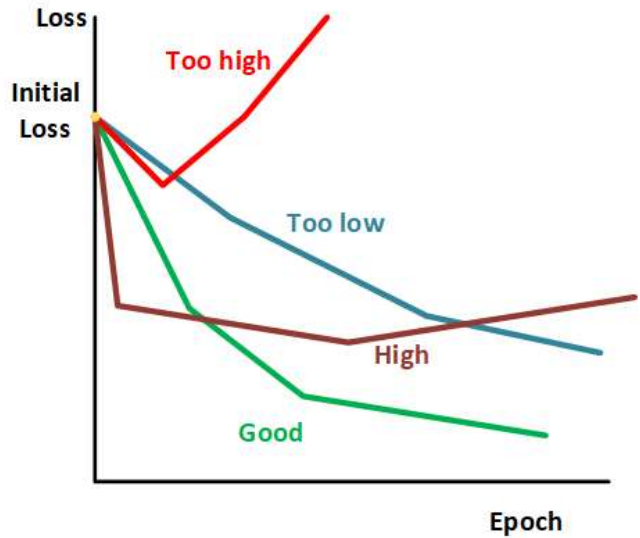

Figure 6. Learning rate graph

In our experiment, a commonly accepted learning rate, 2e-5 (0.00002) is set which is standard for BERT based models, i.e., RoBERTA and DistilBERT. Later, we plot validation vs test accuracy comparison to demonstrate the effectiveness of the selected learning rate.

### C. Fine Tuning

Fine tuning process involves adapting a pre-trained model to get trained on some specific tasks and datasets. This enhances the ability of any pre-trained NLP model to perform any domain-specific task, i.e., email classification for our case. The models are finetuned using training dataset, the 80% of the data which was separated beforehand. Training data is passed in each epoch as batches through the models, calculating the gradient using backpropagation method. To facilitate an efficient batching, DataLoader is used during the training. Code snippet is attached for RoBERTA model in figure 7. A similar approach is employed for DistilBERT as well.

```
train_dataset = EmailDataset(train_df, tokenizer)
val_dataset = EmailDataset(val_df, tokenizer)

train_loader = DataLoader(train_dataset, batch_size=32, shuffle=True)
val_loader = DataLoader(val_dataset, batch_size=64, shuffle=False)

model = RobertaForSequenceClassification.from_pretrained('roberta-base', num_labels=3)

optimizer = AdamW(model.parameters(), lr=2e-5)
num_epochs = 3
```

Figure 7. Code snippet of RoBERTA model DataLoader

TrainLoader is configured for creating mini batches of size, 32, which promotes parallel processing and optimize the memory use. Val_loader is designed to batch of 64 samples for validation ensuring most efficient evaluation method without shuffling the data. RobertaForSequenceClassification class is used to adapt the pre-trained model for specifically email classification task. This class enables an additional classification layer for the target label prediction. The overall fine-tuning process flow diagram is illustrated in figure 8.

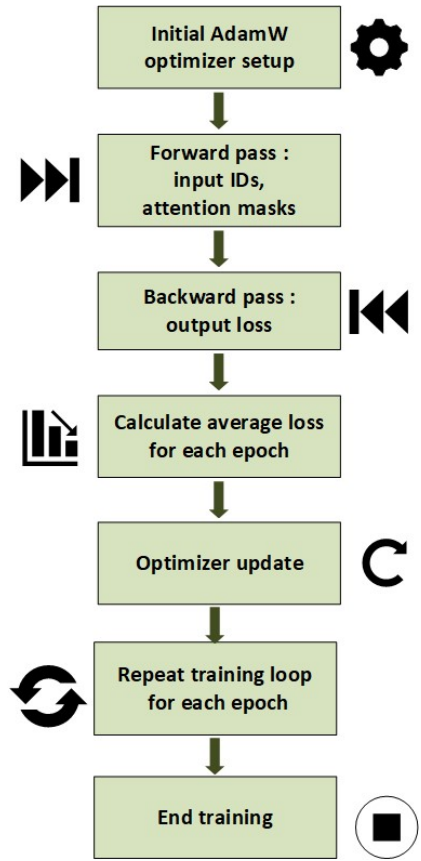

Figure 8. Fine tuning process flow

## 4. Results

The proposed IPSDM model is validated and tested using both unbalanced and balanced datasets. The IPSDM result metrics are compared to baseline modes, i.e., pretrained DistilBERT and RoBERTA models. To assess the performance more comprehensively, various key metrics including overall accuracy, precision, recall and F1-score are calculated. These provide crucial insights of the model performance.

## 4.1 Evaluation metrics

### A. Precision
The ratio of true positive predictions and the total number of positive predictions is called precision. It indicated how many predicted positive samples made by the model are actually positive. The formula for precision is as follow,

$$Precision = \frac{TP}{TP + FP}$$

High precision value suggests that the model's predicted positive instance rate is truly positive and correct. Whereas low precision indicates about making many false positive errors by the model.

### B. Recall
Recall is the measurement of model's sensitivity for understanding true positive rate. It presents the ratio of true positive instances which is predicted as positive by the model. The formula for calculating recall is stated below,

$$Recall = \frac{TP}{TP + FP}$$

Higher recall conveys that the model can successfully predict the positive samples as positive making a little false negative error. However, low recall suggests that a higher number of actual positive samples are getting missed while the model predicts the false negatives.

### C. F1- Score
This is a statistical metric which is the average of precision and recall which balances these values. This provides a comprehensive view on how a model deals with imbalanced datasets by trading off between precision and recall. If either of precision or recall is low, then the overall F1 score will be lower. This metric validates the model's ability for predicting the positive rates and how many instances are actually positive. The formula is as follow,

$$F1\ Score = \frac{2 * Precision * Recall}{Precision + Recall}$$

### D. Accuracy
Accuracy is the ratio of accurately predicted samples to the total number of samples made by the model[38]. It is calculated by following formula,

$$Accuracy = \frac{TP + TN}{Total\ samples}$$

However, some notable points need to consider carefully when interpreting the model accuracy because it suffers from some limitations while dealing with imbalanced data. Feature distribution across all the classes is required to be observed meticulously. Otherwise, it might raise a biased classification result. Hence, in this study, all of the essential metrics are calculated and combined together to interpret our proposed IPSDM results after running a vigilant examination.

## 4.2 Imbalanced Dataset Results
This experiment was initially carried on imbalanced datasets to assess IPSDM model's performance on imbalanced dataset. The initial collected dataset was highly imbalanced having a majority class, ham (Figure 2). Comparison tables (Table 1 and 2) and graphs (Figure 9 and 10) between baseline model's performance and IPSDM model's performance clearly reflect that IPSDM has a better performance in the imbalanced setting. Although due to highly uneven distribution of data samples across the three classes, the model performance is a biased towards 'ham' class, still it has achieved comparatively higher values than the baseline models.

| Evaluation Metrics | Base DistilBERT | IPSDM |
|---|---|---|
| Validation Accuracy | 30.28% | 51.32% |
| Test Accuracy | 31.60% | 53.67% |
| Validation Precision | 0.841 | 0.972 |
| Test Precision | 0.852 | 0.981 |
| Validation Recall | 0.302 | 0.561 |
| Test Recall | 0.311 | 0.582 |
| Validation F1-Score | 0.432 | 0.613 |
| Test F1-Score | 0.451 | 0.621 |

Table 1. Baseline DistilBERT vs IPSDM performance (imbalanced dataset)

In Figures 9 and 10, the precision values are higher than recall for both cases (DistilBERT and RoBERTA). In the context of highly imbalanced characteristics of this dataset, the model can identify the majority class, 'ham', however, for the model struggles for classifying the minor classes, 'spam' and 'phishing'.

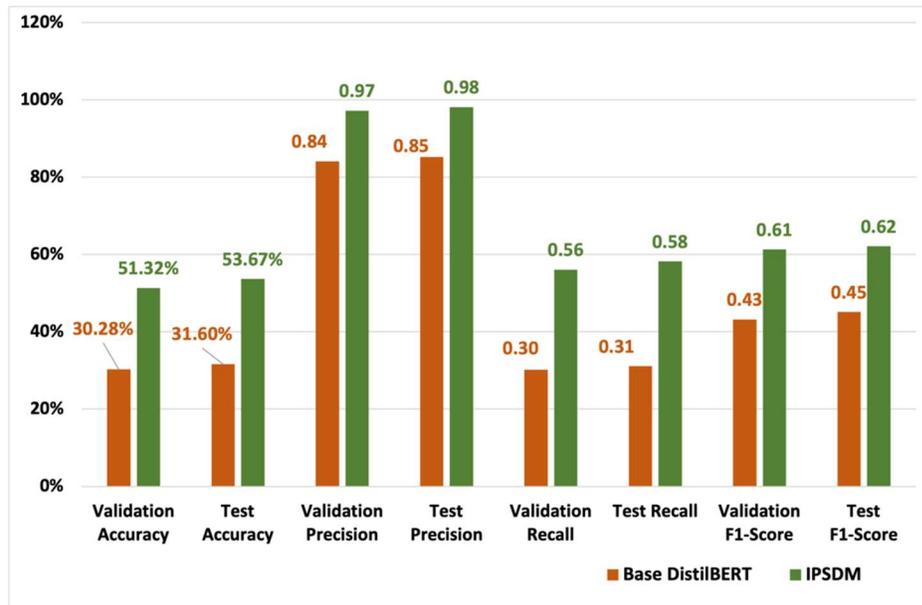

Figure 9. Comparison graph of baseline DistilBERT vs IPSDM performance (imbalanced dataset)

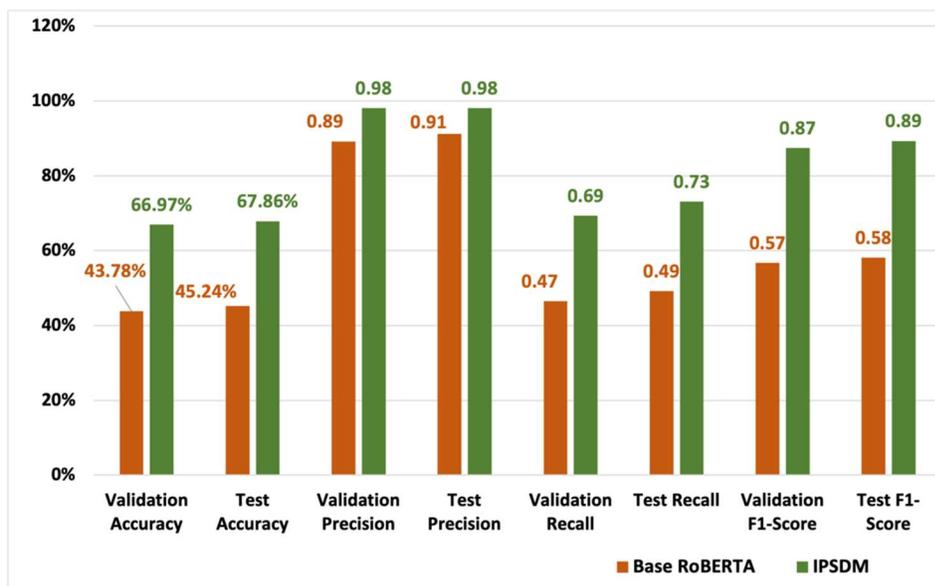

Figure 10. Comparison graph of baseline RoBERTA vs IPSDM performance (imbalanced dataset)

There is a noticeable disparity between Precision and Recall values for both models. Recall values are considerably lower compared to the precision. Validation and test recall for base DistilBERT model are 0.30 and 0.31 (shown in Table 1). For base RoBERTA model, the recall values are 0.47 and 0.49 (shown in Table 2) which suggest that the models are facing challenges for identifying the minor classes, 'spam' and 'phishing' due to the imbalanced nature. However, it is noteworthy that the performance of IPSDM for both DistilBERT and RoBERTA is notably higher even the dataset is imbalanced.

| Evaluation Metrics | Base RoBERTA | IPSDM |
|---|---|---|
| Validation Accuracy | 43.78% | 66.97% |
| Test Accuracy | 45.24% | 67.86% |
| Validation Precision | 0.892 | 0.981 |
| Test Precision | 0.912 | 0.981 |
| Validation Recall | 0.465 | 0.693 |
| Test Recall | 0.492 | 0.731 |
| Validation F1-Score | 0.567 | 0.874 |
| Test F1-Score | 0.581 | 0.893 |

Table 2. Baseline RoBERTA vs IPSDM performance (imbalanced dataset)

## 4.3 Balanced Dataset Results

The collected email datasets have been resampled and balanced. After preparing this balanced dataset, baseline DistilBERT and RoBERTA models were trained and validated. Again, using the similar dataset, we worked on model optimization and fine tuning. The evaluation metrics of our proposed model, IPSDM and the baseline models are tabulated in Tables 3 and 4. Accuracy, precision, recall for both validation and test cases are presented here. Also, the values are illustrated in comparison graphs (Figure 11 and Figure 12).

| Evaluation Metrics | Base DistilBERT | IPSDM |
|---|---|---|
| Validation Accuracy | 82.63% | 97.50% |
| Test Accuracy | 88.95% | 97.10% |
| Validation Precision | 0.8543 | 0.9755 |
| Test Precision | 0.9025 | 0.9716 |
| Validation Recall | 0.6971 | 0.9750 |
| Test Recall | 0.7532 | 0.9710 |
| Validation F1-Score | 0.8867 | 0.9749 |
| Test F1-Score | 0.8943 | 0.9710 |

Table 3. Baseline DistilBERT vs IPSDM performance (balanced dataset)

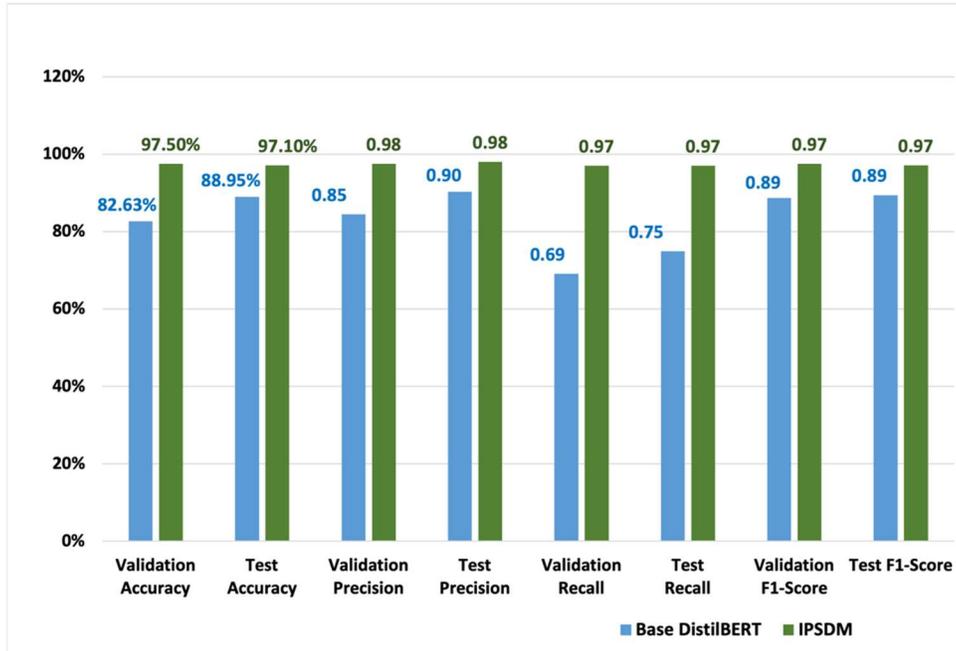

Figure 11. Comparison graph of Baseline DistilBERT vs IPSDM performance (balanced dataset)

The evaluation metrics from Figures 11 and 12 exhibit an increase in validation accuracy- approximately 14.87% and 11.89%; test accuracy approximately 8.15% and 5.71% respectively for base DistilBERT and RoBERTA models vs IPSDM. A consistent rise in F1scores suggests that the IPSDM has elevated performance across both cases. This score is the harmonic mean of recall and precision which is a crucial metric for assessing the balance between the crucial aspects of classification performance.

| Evaluation Metrics | Base ROBERTA | IPSDM |
| --- | --- | --- |
| Validation Accuracy | 87.10% | 98.99% |
| Test Accuracy | 93.29% | 99.00% |
| Validation Precision | 0.921 | 0.982 |
| Test Precision | 0.853 | 0.991 |
| Validation Recall | 0.903 | 0.989 |
| Test Recall | 0.923 | 0.991 |
| Validation F1-Score | 0.911 | 0.982 |
| Test F1-Score | 0.931 | 0.985 |

Table 4. Baseline ROBERTA vs IPSDM performance (balanced dataset)

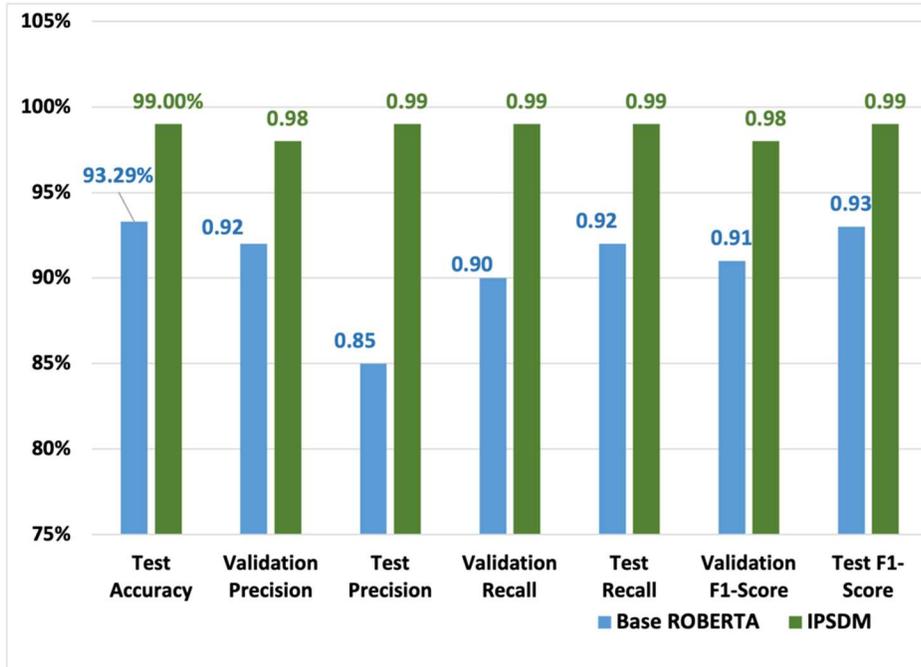

Figure 12. Comparison graph of Baseline RoBERTA vs IPSDM performance (balanced dataset)

## 4.5 Avoiding Overfitting

A common issue in statistical modellings and machine learning is overfitting which occurs when a model is performs too well on the training dataset, however, too poorly on the new or unseen data, i.e., testing dataset. Overfitting can be effectively managed in balanced situations while a model has consistent performance on validation and test datasets. A close alignment between test and validation accuracy suggests that the classification models yield good results on unseen, new data. In the balanced scenario, test and validation accuracy values indicate minimal disparity, i.e., 97.10% vs 97.50% and 99. 00% vs 98.99%.

| Model Name | Validation accuracy | Test accuracy |
|---|---|---|
| Balanced_ IPSDM/ DistilBERT | 97.50% | 97.10% |
| Balanced_ IPSDM/ RoBERTA | 98.99% | 99.00% |
| Imbalanced_ IPSDM/ DistilBERT | 51.32% | 53.67% |
| Imbalanced_ IPSDM/ RoBERTA | 66.97% | 67.86% |

Table 5. Validation vs test accuracy

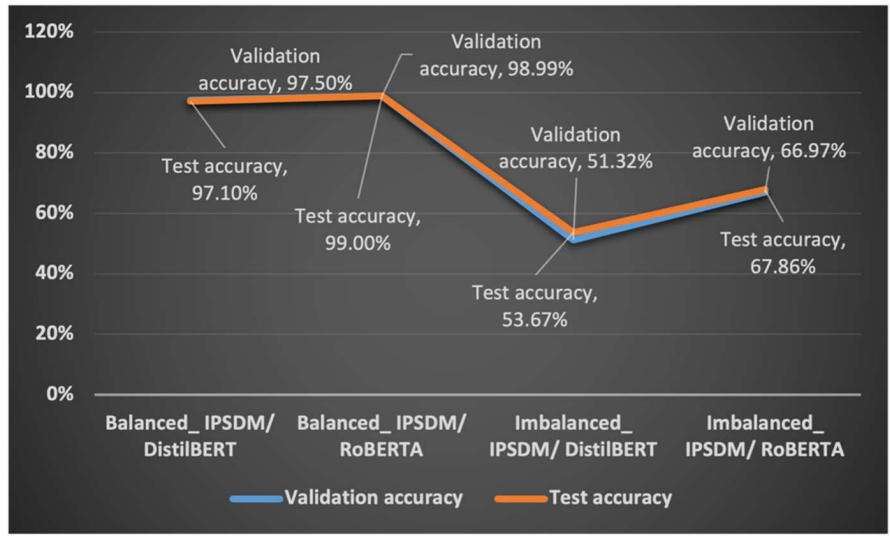

Figure 13. Validation vs test accuracy graph

Based on the Table 5 and Figure 13 data, there is not a large gap between validation and test accuracy. When training or validation accuracy is notably higher than test accuracy, there is a high chance of overfitting. Moreover, the precision, recall and F1 measures from table 1 through 4 also suggest a harmonic distribution of these metrics which is also a positive indication. The comparison graph for both validation and test accuracy lines are almost overlapping with each other indicating that the model is performing well on the unseen data.

## 5. Discussion

The results from both imbalanced and balanced settings depict an enhancement in performance for IPSDM model. Validation and test accuracy are separately measured to understand if there is any overfitting issue persist. Baseline DistilBERT has 82.63% of validation accuracy and 88.95% of testing accuracy whereas IPSD has 97.50% and 97.10% validation and test accuracy respectively. The baseline model's accuracy variation is 6 (+-32) % in training and testing performance reveal that base DistilBERT is exhibiting a minor overfitting problem. However, this has been effectively handled during the development of IPSDM DistilBERT version. Again, a similar trait is visible for base RoBERTA and IPSDM for RoBERTA as well. Base RoBERTA mpdel's validation and testing accuracy gap is around 6 (+- 19) % whereas IPSDM has 0.01% of difference between these two values.

Such evaluation has been also extended to imbalanced dataset to analyze how IPSDM is performing in challenging scenario. In the imbalanced setting, (Table 3 and 4) convey that our proposed model has outperformed the baseline models in this scenario as well. Due to the heavy skewness of sample distribution, the results are biased towards 'ham' class. The precision values for both baseline and IPSDM models are notably higher. 0.85 and 0.98 for base DistilBERT and IPSDM test precision; 0.91 and 0.98 for base RoBERTA and IPSDM respectively present that the model is predicting most of the instances as 'ham'. This impact results in higher precision and lower recall. However, later applying ADASYN, an advanced sampling technique, this class imbalanced situation is handled at the initial stage. A prominent change in performance is hence demonstrated in IPSDM models both for DistilBERT and RoBERTA both for balanced and imbalanced datasets scenarios. Our proposed such methodological approach for enhancing training performance can be further extended to any BERT based LLM architecture.

## 6. Conclusion and Future Directions

Solving long standing societal issues via radical new approaches, specifically LLMs, shows great promise to improving the lives and experiences of computing users the world over. Phishing and Spam have long since been an issue causing lost time and straining financial resources of consumers and organizations. We demonstrate how leveraging new technology can be applied to these persistent challenges. LLMs offer society great benefits and we have only scratched the surface on their potential. In the future, improving the quality of life via multiple dimensions will be realized such as medical diagnoses, chat-bots, education, and security to name a few. This work demonstrates how LLMs can be leveraged to detect phishing and spam by leveraging LLMs and then presenting our fine-tuned version, IPSDM. Following the proposed mechanism, modified DistilBERT could achieve 97.50% of validation and 97.10% of test accuracy with a F1-score of 0.97. Again, the modified RoBERTA model obtained 98.99% of validation and 99.00% of test accuracy including a F1-score of 0.98. The result of this study presents the effectiveness of IPSDM model while reducing the overfitting issues and handling imbalanced datasets. The attained accuracy has surpassed the existing state-of-the art models.

Future work entails further refinement of IPSDM via incorporation of additional tuning techniques as well as hyper-parameter tuning and combining with ensemble modeling. Applying data augmentation such as text rotation, contrastive learning, synonym replacement might also assist in increasing the diversity and improving the training performance. Furthermore, the field of Large Language Models has attracted substantial investment from industry and consumers causing it to develop rapidly with new open-source models being released nearly daily. We aim to experiment with further LLMs such as Meta's Llama and Llama 2. Infusing such solutions into chatbot, web applications and other real-world practical systems would serve society in numerous valuable ways.